\documentclass[10pt,twocolumn,letterpaper]{article}

\usepackage{cvpr}
\usepackage{times}
\usepackage{epsfig}
\usepackage{graphicx}
\usepackage{amsmath}
\usepackage{amssymb}
\usepackage{subfigure}
\usepackage{multirow}

\usepackage[pagebackref=true,breaklinks=true,pdftex,letterpaper=true,colorlinks,citecolor=blue,bookmarks=false]{hyperref}

\cvprfinalcopy 

\def\cvprPaperID{4616} 

\def \algfullname{\emph{task-aware spatial disentanglement}}
\def \algname{TSD}
\def \loss{PC}
\ifcvprfinal\fi
\begin{document}

\title{Revisiting the Sibling Head in Object Detector}

\author{Guanglu Song\textsuperscript{\rm 1}\ \ \ \ \  Yu Liu\textsuperscript{\rm 2}\thanks{Corresponding author}\ \ \ \ \ \  Xiaogang Wang\textsuperscript{\rm 2} \\
\textsuperscript{\rm 1}SenseTime X-Lab\\
\textsuperscript{\rm 2}The Chinese University of Hong Kong, Hong Kong\\
\textsuperscript{\rm 1}songguanglu@sensetime.com, \textsuperscript{\rm 2}\{yuliu, xgwang\}@ee.cuhk.edu.hk
}

\maketitle

\begin{abstract}
   The ``shared head for classification and localization'' (sibling head), firstly denominated in Fast RCNN~\cite{girshick2015fast}, has been leading the fashion of the object detection community in the past five years. 
   This paper provides the observation that the spatial misalignment between the two object functions in the sibling head can considerably hurt the training process, but this misalignment can be resolved by a very simple operator called \algfullname{}~(\algname{}). Considering the classification and regression, \algname{} decouples them from the spatial dimension by generating two disentangled proposals for them, which are estimated by the shared proposal. This is inspired by the natural insight that for one instance, the features in some salient area may have rich information for classification while these around the boundary may be good at bounding box regression. Surprisingly, this simple design can boost all backbones and models on both MS COCO and Google OpenImage consistently by $\sim$3\% mAP. Further, we propose a progressive constraint to enlarge the performance margin between the disentangled and the shared proposals, and gain $\sim$1\% more mAP. We show the \algname{} breaks through the upper bound of nowadays single-model detector by a large margin (\textbf{mAP 49.4 with ResNet-101, 51.2 with SENet154}), and is the core model of our \textbf{1st place solution on the Google OpenImage Challenge 2019}.

\end{abstract}

\begin{figure}[t]
\centering
\includegraphics[width=1.0\linewidth]{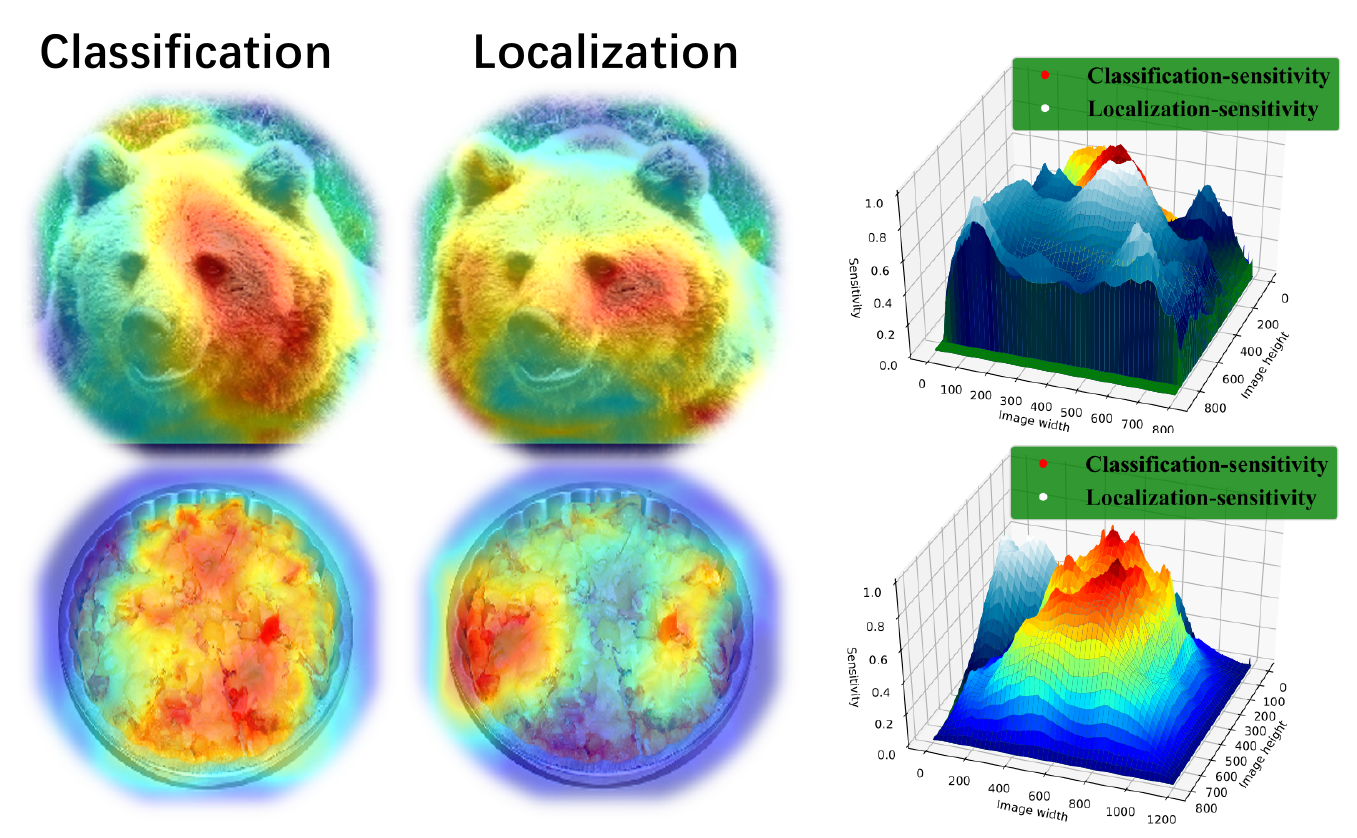}
   \caption{Illustration of the task spatial misalignment. The first column is the sensitive location for classification and the second column is the sensitive location for localization. The third column is the 3D visualization of the sensitivity distribution.}
\label{fig:moti}
\end{figure}

\section{Introduction}
Since the breakthrough of object detection performance has been achieved by seminal R-CNN families~\cite{girshick2015region,girshick2015fast,ren2015faster} and powerful FPN~\cite{lin2017feature}, the subsequent performance enhancement of this task seems to be hindered by some concealed bottlenecks.
Even the advanced algorithms bolstered by AutoML~\cite{ghiasi2019fpn,Xu_2019_ICCV} have been delved, the performance gain is still limited to an easily accessible improvement range.
As the most obvious distinction from the generic object classification task, the specialized sibling head for both classification and localization comes into focus and is widely used in most of advanced detectors including single stage family~\cite{liu2016ssd,song2018beyond,hao2017scale}, two-stage family~\cite{dai2017deformable,li2019gradient,zeng2017crafting,liu2017recurrent,li2019zoom} and anchor free family~\cite{law2018cornernet}.
Considering the two different tasks share almost the same parameters, a few works become conscious about the conflict between the two object functions in the sibling head and try to find a trade-off way.

IoU-Net~\cite{jiang2018acquisition} is the first to reveal this problem. They find the feature which generates a good classification score always predicts a coarse bounding box.
To handle this problem, they first introduce an extra head to predict the IoU as the localization confidence, and then aggregate the localization confidence and the classification confidence together to be the final classification score. This approach does reduce the misalignment problem but in a compromise manner -- the essential philosophy behind it is relatively raising the confidence score of a tight bounding box and reduce the score of a bad one. The misalignment still exists in each spatial point. 
Along with this direction, Double-Head R-CNN~\cite{wu2019rethinking} is proposed to disentangle the sibling head into two specific branches for classification and localization, respectively. Despite of elaborate design of each branch, it can be deemed to disentangle the information by adding a new branch, essentially reduce the shared parameters of the two tasks.
Although the satisfactory performance can be obtained by this detection head disentanglement, conflict between the two tasks still remain since the features fed into the two branches are produced by ROI Pooling from the same proposal.

In this paper, we meticulously revisit the sibling head in the anchor-based object detector to seek the essence of the tasks misalignment.
We explore the spatial sensitivity of classification and localization on the output feature maps of each layer in the feature pyramid of FPN. Based on the commonly used sibling head (a fully connected head \emph{2-fc}), we illustrate the spatial sensitive heatmap in Figure.\ref{fig:moti}.
The first column is the spatial sensitive heatmap for classification and the second column is for localization. The warmer the better for the color. We also show their 3D visualizations in the third column.
It's obvious that for one instance, the features in some salient areas may have rich information for classification while these around the boundary may be good at bounding box regression.
This essential tasks misalignment in spatial dimension greatly limits the performance gain whether evolving the backbone or enhancing the detection head. In other words, if a detector try to infer the classification score and regression result from a same spatial point/anchor, it will always get an imperfect trade-off result.

This significant observation motivates us to rethink the architecture of the sibling head. The optimal solution for the misalignment problem should be explored by the spatial disentanglement. Based on this, we propose a novel operator called \algfullname{} (\algname{}) to resolve this barrier.
The goal of \algname{} is to \textbf{spatially disentangle the gradient flows of classification and localization}. To achieve this, \algname{} generates two disentangled proposals for these two tasks, based on the original proposal in classical sibling head.
It allows two tasks to adaptively seek the optimal location in space without compromising each other.
With the simple design, the performance of all backbones and models on both MS COCO and Google OpenImage are boosted by $\sim$3\% mAP. 
Furthermore, we propose a \emph{progressive constraint} (\loss{}) to enlarge the performance margin between \algname{} and the classical sibling head.
It introduces the hyper-parameter \emph{margin} to advocate the more confident classification and precise regression. $\sim$1\% more mAP is gained on the basis of \algname{}.
Whether for variant backbones or different detection frameworks, 
the integrated algorithms can steadily improve the performance by $\sim$4\% and even $\sim$6\% for lightweight MobileNetV2.
Behind the outstanding performance gains, only a slight increased parameter is required, which is negligible for some heavy backbones.

To summarize, the contributions of this paper are as follows:

1) We delve into the essential barriers behind the tangled tasks in RoI-based detectors and reveal the bottlenecks that limit the upper bound of detection performance.

2) We propose a simple operator called \algfullname{} (\algname{}) to deal with the tangled tasks conflict. Through the task-aware proposal estimation and the detection head, it could generate the task-specific feature representation to eliminate the compromises between classification and localization.

3) We further propose a \emph{progressive constraint} (\loss{}) to enlarge the performance margin between \algname{} and the classical sibling head.

4) We validate the effectiveness of our approach on the standard COCO benchmark and large-scale OpenImage dataset with thorough ablation studies. 
Compared with the state-of-the-art methods,
our proposed method achieves the \textbf{mAP of 49.4 using 
a single model with ResNet-101 backbone and mAP of 51.2 with heavy SENet154.}



\begin{figure*}[t!]
\centering
\includegraphics[width=.9\linewidth]{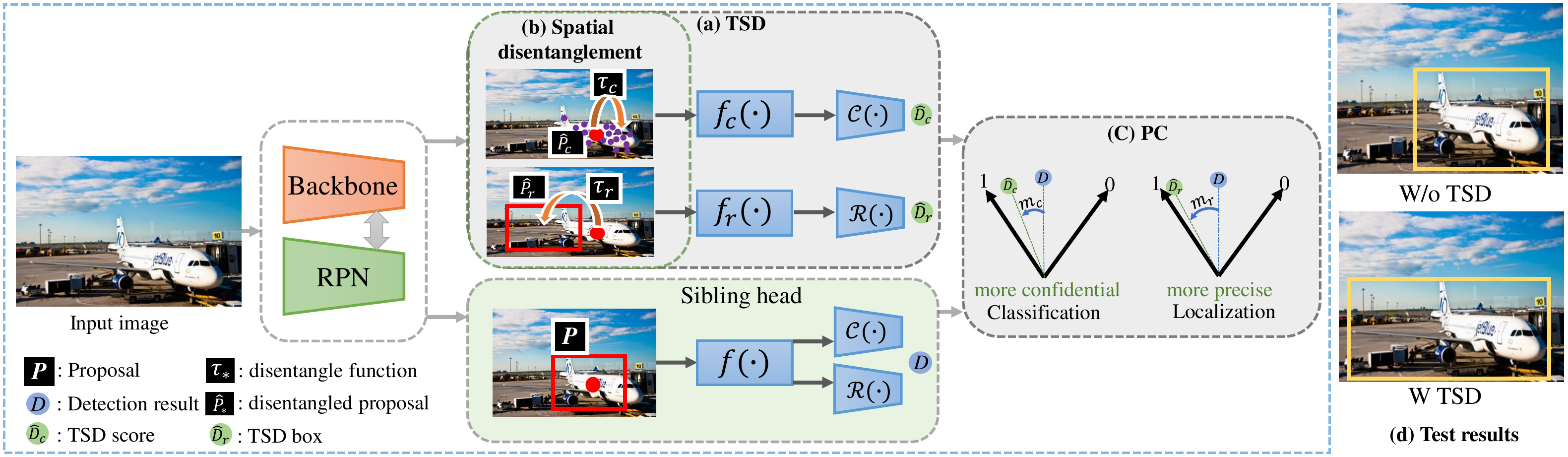}
   \caption{Illustration of the proposed \algname{} cooperated with Faster RCNN~\cite{ren2015faster}. Input images are first fed into the FPN backbone and then, region proposal $P$ is generated by RPN. \algname{} adopts the RoI feature of $P$ as input and estimates the derived proposals $\hat{P}_c$ and $\hat{P}_r$ for classification and localization. Finally, two parallel branches are used to predict specific category and regress precise box, respectively.}
\label{fig:intro}
\end{figure*}



\section{Methods}
In this section, we first describe the overall framework of our proposed \algfullname{} (\algname{}), then detail the sub-modules in Sec.~\ref{L1} and ~\ref{L2}. Finally, we delve into the inherent problem in sibling head and demonstrate the advantage of \algname{}.

\subsection{\algname}
As shown in Figure.\ref{fig:intro}~(a),
denote a rectangular bounding box proposal as $P$ and the ground-truth bounding box as $\mathcal{B}$ with class ${y}$,
the classical Faster RCNN~\cite{ren2015faster} aims to minimize the classification loss and localization loss based on the shared $P$:
\begin{equation}
\begin{split}
\mathcal{L} = \mathcal{L}_{cls}(\mathcal{H}_1 (F_l, P), y) + \mathcal{L}_{loc}(\mathcal{H}_2(F_l, P), \mathcal{B}) \label{L3}
\end{split}
\end{equation}
where $\mathcal{H}_1(\cdot)$ = $\{f(\cdot), \mathcal{C}(\cdot)\}$ and $\mathcal{H}_2(\cdot)$ = $\{f(\cdot), \mathcal{R}(\cdot)\}$.
$f(\cdot)$ is the feature extractor and $\mathcal{C}(\cdot)$ and $\mathcal{R}(\cdot)$ are the functions for transforming feature to predict specific category and localize object.
Seminal work~\cite{wu2019rethinking} thinks the shared $f$ for classification and localization is not optimal, and they disentangle it to $f_c$ and $f_r$ for classification and regression, respectively. Although the appropriate head-decoupling brings a reasonable improvement, the inherent conflict caused by the tangled tasks in the spatial dimension is still lurking.

For this potential problem, our goal is to alleviate the inherent conflict in sibling head by disentangling the tasks from the spatial dimension. 
We propose a novel \algname{} head for this goal as shown in Figure~\ref{fig:intro}. 
In \algname{}, the Eq.~\ref{L3} can be written as:
\begin{equation}
\begin{split}
\mathcal{L} = \mathcal{L}^D_{cls}(\mathcal{H}^D_1 (F_l, \hat{P}_c), y) + \mathcal{L}^D_{loc}(\mathcal{H}^D_2(F_l, \hat{P}_r), \mathcal{B})
\end{split}
\end{equation}
where disentangled proposals $\hat{P}_c=\tau_c(P,\Delta C)$ and $\hat{P}_r = \tau_r(P,\Delta R)$ are estimated from the shared $P$.
$\Delta C$ is a pointwise deformation of $P$ and $\Delta R$ is a proposal-wise translation.
In \algname{}, $\mathcal{H}^D_1(\cdot)$ = $\{f_c(\cdot), \mathcal{C}(\cdot)\}$ and $\mathcal{H}^D_2(\cdot)$ = $\{f_r(\cdot), \mathcal{R}(\cdot)\}$.

In particular, \algname{} tasks the RoI feature of $P$ as input, and then generates the
disentangled proposals $\hat{P}_c$ and $\hat{P}_r$ for classification and localization, respectively.
Different tasks can be disentangled from the spatial dimension via the separated proposals.
The classification-specific feature maps $\hat{F}_c$ and localization-specific feature maps $\hat{F}_r$ can be generated through parallel branches. 
In the first branch, $\hat{F}_c$ is fed into a three-layer fully connected networks for classification.
In the second branch, the RoI feature $\hat{F}_r$ corresponding to derived proposal $\hat{P}_r$ will be extracted and fed into a similar architecture with the first branch to perform localization task. By disentangling the shared proposal for the classification and localization, \algname{} can learn the task-aware feature representation adaptively. 
\algname{} is applicable to most existing RoI-based detectors.
As the training procedure adopts an end-to-end manner cooperated with the well-designed progressive constraint (\loss{}), it is robust to the change of backbones and input distributions (e.g., training with different datasets.).

\subsection{Task-aware spatial disentanglement learning}\label{L1}
Inspired by Figure.\ref{fig:moti}, we introduce the \algfullname{} learning to alleviate the misalignment caused by the shared spatial clues.
As shown in Figure.\ref{fig:intro}~(b), define the RoI feature of $P$ as $F$, we embed the deformation-learning manner into \algname{} to achieve this goal.
For localization, a three-layer fully connected network $\mathcal{F}_r$ is designed to generate a proposal-wise translation on $P$ to produce a new derived proposal $\hat{P}_r$. This procedure can be formulated as:
\begin{equation}
\begin{split}
\Delta R = \gamma\mathcal{F}_r (F;\theta_r) \cdot (w, h)
\end{split}
\end{equation}
where $\Delta R\in \mathbb{R}^{1\times 1 \times 2}$ and the output of $\mathcal{F}_r$ for each layer is \{256, 256, 2\}.
$\gamma$ is a pre-defined scalar to modulate the magnitude of the $\Delta R$ and (w, h) is the width and height of $P$.
The derived function $\tau_r(\cdot)$ for generating $\hat{P}_r$ is:
\begin{equation}
\begin{split}
\hat{P}_r  = P + \Delta R \label{E4}
\end{split}
\end{equation}
Eq.~\ref{E4} indicates the proposal-wise translation where the coordinate of each pixel in $P$ will be translated to a new coordinate with the same $\Delta R$.
The derived proposal $\hat{P}_r$ only focuses on the localization task and in the pooling function, we adopt the bilinear interpolation the same as~\cite{dai2017deformable} to make $\Delta R$ differentiable.

For classification, given the shared $P$,
a pointwise deformation on a regular grid $k\times k$ is generated to estimate a derived proposal $\hat{P}_c$ with an irregular shape.
For (x,y)-th grid, the translation $\Delta C(x,y,*)$ is performed on the sample points in it to obtain the new sample points for $\hat{P}_c$.
This procedure can be formulated as:
\begin{equation}
\begin{split}
\Delta C = \gamma \mathcal{F}_c (F;\theta_c) \cdot (w, h)
\end{split}
\end{equation}
where $\Delta C\in \mathbb{R}^{k\times k \times 2}$. $\mathcal{F}_c$ is a three-layer fully connected network with output \{256, 256, $k\times k\times 2$\} for each layer and $\theta_c$ is the learned parameter. The first layer in $\mathcal{F}_r$ and $\mathcal{F}_c$ is shared to reduce the parameter.
For generating feature map $\hat{F}_c$ by irregular $\hat{P}_c$, we adopt the same operation with deformable RoI pooling~\cite{dai2017deformable}:
\begin{equation}
\begin{split}
\hat{F}_c(x,y) \!=\! \sum_{p\in G(x,y) }\! \frac{\mathcal{F}_{B}(p_0+\Delta C(x,y,1), p_1+\Delta C(x,y,2))}{|G(x,y)|}
\end{split}
\end{equation}
where $G(x,y)$ is the (x,y)-th grid and $|G(x,y)|$ is the number of sample points in it.
$(p_x, p_y)$ is the coordinate of the sample point in grid $G(x,y)$ and $\mathcal{F}_{B}(\cdot)$ is the bilinear interpolation~\cite{dai2017deformable} to make the $\Delta C$ differentiable.

\subsection{Progressive constraint}\label{L2}
At the training stage, the \algname{} and the sibling detection head defined in Eq.~\ref{L3} can be jointly optimized by $\mathcal{L}_{cls}$ and $\mathcal{L}_{loc}$.
Beyond this, we further design the \emph{progressive constraint} (\loss{}) to improve the performance of \algname{} as shown in Figure.\ref{fig:intro}~(c).
For classification branch, \loss{} is formulated as:
\begin{equation}
\begin{split}
\mathcal{M}_{cls} \!=\! |  \mathcal{H}_1(y | F_l, P) \!-\! \mathcal{H}^D_1(y| F_l, \tau_c(P,\Delta C)) \!+\! m_c |_+
 \end{split}
\end{equation}
where $\mathcal{H}(y|\cdot)$ indicates the confidence score of the $y$-th class and $m_c$ is the predefined margin. $|\cdot|_+$ is same as ReLU function.
Similarly, for localization, there are:
\begin{equation}
\begin{split}
\mathcal{M}_{loc} = |  IoU(\hat{\mathcal{B}}, \mathcal{B}) - IoU(\hat{\mathcal{B}}_D, \mathcal{B}) + m_r |_+
 \end{split}
\end{equation}
where $\hat{\mathcal{B}}$ is the predicted box by sibling head and $\hat{\mathcal{B}}_D$ is regressed by $\mathcal{H}^D_2(F_l, \tau_r(P, \Delta R))$.
If $P$ is a negative proposal, $M_{loc}$ is ignored.
According to these designs, the whole loss function of \algname{} with Faster RCNN can be define as:
\begin{equation}
\begin{split}
\mathcal{L}\!=\! \underbrace{\mathcal{L}_{rpn} \!+\! \mathcal{L}_{cls} \!+\! \mathcal{L}_{loc}  }_{classical\; loss}
+  \underbrace{  \mathcal{L}^D_{cls} \!+\! \mathcal{L}^D_{loc} \!+\! \mathcal{M}_{cls} \!+\! \mathcal{M}_{loc} }_{\algname{}\; loss}
 \end{split}
\end{equation}
We directly set the loss weight to 1 without carefully adjusting it.
Under the optimization of $\mathcal{L}$, \algname{} can adaptively learn the task-specific feature representation for classification and localization, respectively.
Extensive experiments in Sec.\ref{exp} indicates that disentangling the tangled tasks from the spatial dimension can significantly improve the performance.

\subsection{Discussion in context of related works}
In this section, we delve into the inherent conflict in tangled tasks.
Our work is related to previous works in different aspects. We discuss the relations and differences in detail.
\subsubsection{Conflict in sibling head with tangled tasks}\label{discuss}

Two core designs in classical Faster RCNN are predicting the category for a given proposal and learning a regression function.
Due to the essential differences in optimization, classification task requires translation-agnostic property and to the contrary, localization task desires translation-aware property.
The specific translation sensitivity property for classification and localization can be formulated as:
\begin{equation}
\begin{split}
\mathcal{C}(f(F_l, P)) = \mathcal{C}(f(F_l, P + \varepsilon)), \\ 
\mathcal{R}(f(F_l, P)) \neq \mathcal{R}(f(F_l, P + \varepsilon))  
 \end{split}
\end{equation}
where $\forall \varepsilon, IoU(P + \varepsilon, \mathcal{B}) \geq T$. $\mathcal{C}$ is to predict category probability and $\mathcal{R}$ is the regression function whose output is $(\Delta\hat{x}, \Delta\hat{y}, \Delta\hat{w}, \Delta\hat{h})$.
$f(\cdot)$ is the shared feature extractor in classical sibling head and $T$ is the threshold to determine whether $P$ is a positive sample.
There are entirely different properties in these two tasks.
The shared spatial clues in $F_l$ and feature extractor for these two tasks will become the obstacles to hinder the learning.
Different from ~\cite{wu2019rethinking,jiang2018acquisition,dai2017deformable,zhu2019deformable} where the evolved backbone or feature extractor is designed, \algname{} decouples the classification and regression from spatial dimension by separated $\hat{P}_*$ and $f_*(\cdot)$.

\subsubsection{Different from other methods}
IoU-Net~\cite{jiang2018acquisition} first illustrates the misalignment between classification and regression. To alleviate this, it directly predicts the IoU to adjust the classification confidence via an extra branch.
Unfortunately, this approach does not solve the inherent conflict between tangled tasks.
For this same problem, Double-Head R-CNN~\cite{wu2019rethinking} explores the optimal architectures for classification and localization, respectively.
To learn more effective feature representation, DCN~\cite{dai2017deformable} with deformable RoI pooling is proposed to extract the semantic information from the irregular region.
Whether evolving the backbone or adjusting the detection head,
performance can be improved, but the increase is limited.

In this paper, we observe that the essential problem behind the limited performance is the misaligned sensitivity in the spatial dimension between classification and localization.
Neither designing better feature extraction methods nor searching for the best architecture can solve this problem. 
In this dilemma, \algname{} is proposed to decouple the classification and localization from both the spatial dimension and feature extractor.
\algname{} first performs spatial disentanglement for classification and localization via separated proposals and feature extractors to break the predicament. With the further well-designed \loss{}, it can learn the optimal sensitive location for classification and localization, respectively.
Moreover, \algname{} is still applicable to DCN~\cite{dai2017deformable} although deformable RoI pooling in DCN is used to assist in estimating $\hat{F}_c$.
By \algfullname{}, the simple \algname{} can easily achieve excellent performance for different backbones.

\section{Experiments}\label{exp}
We perform extensive experiments with variant backbones on the 80-category MS-COCO dataset~\cite{lin2014microsoft} (object detection and instance segmentation) and 500-category OpenImageV5 challenge dataset~\cite{OpenImages}. 
For COCO dataset, following the standard protocol~\cite{lu2019grid}, training is performed on the union of 80k \textit{train} images and 35k subset of \textit{val} images and testing is evaluated on the remaining 5k val images (\textit{minival}). We also report results on 20k \textit{test-dev}.
For OpenImage dataset, following the official protocol~\cite{OpenImages}, the model is trained on 1,674,979 training images and evaluated on the 34,917 val images.
The AP$_{.5}$ on public leaderboard is also reported.

\subsection{Implementation details}
We initialize weights from pre-trained models on ImageNet~\cite{russakovsky2015imagenet} and the configuration of hyper-parameters follows existing Faster RCNN~\cite{ren2015faster}.
Images are resized such that the shorter edge is 800 pixels.
The anchor scale and aspect ratio are set to 8 and \{0.5, 1, 2\}.
We train the models on 16 GPUs (effective mini-batch size is 32) for 13 epochs, with a learning rate warmup strategy~\cite{goyal2017accurate} from 0.00125 to 0.04 in the first epoch. We decrease the learning rate by 10 at epoch 8 and epoch 11, respectively.
RoIAlign~\cite{he2017mask} is adopted in all experiments, and the pooling size is 7 in both $\mathcal{H}^*_1$ and $\mathcal{H}^*_2$.
We use SGD to optimize the training loss with 0.9 momentum and 0.0001 weight decay. No data augmentations except standard horizontal flipping are used.
Synchronized BatchNorm mechanism~\cite{peng2018megdet,goyal2017accurate} is used to make multi-GPU training more stable.
At the inference stage, NMS with 0.5 IoU threshold is applied to remove duplicate boxes.
For experiments in the OpenImage dataset, class-aware sampling is used.

\subsection{Ablation studies}
In this section, we conduct detailed ablation studies on COCO $\emph{minival}$ to evaluate the effectiveness of each module and illustrate the advance and generalization of the proposed \algname{}. $m_c$ and $m_r$ are set to 0.2 in these experiments.

\begin{figure}[t]
  \centering 
  \subfigure[D$_{s8}$]{ 
    \label{fig:subfig:a} 
    \includegraphics[width=1.5in]{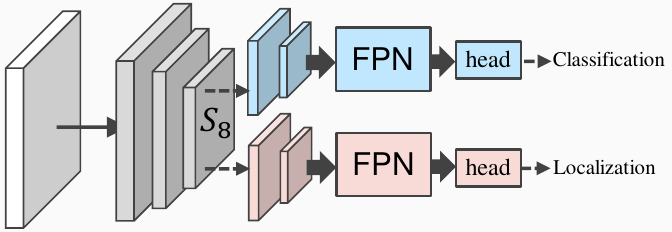}} 
  \subfigure[D$_{s16}$]{ 
    \label{fig:subfig:b} 
    \includegraphics[width=1.5in]{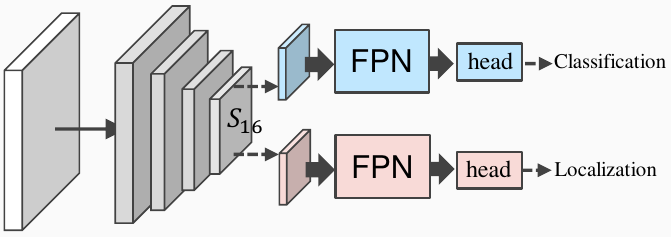}}
\subfigure[D$_{s32}$]{ 
    \label{fig:subfig:c} 
    \includegraphics[width=1.5in]{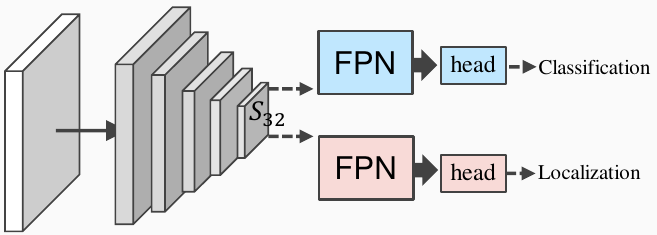}} 
\subfigure[D$_{head}$]{ 
    \label{fig:subfig:d} 
    \includegraphics[width=1.5in]{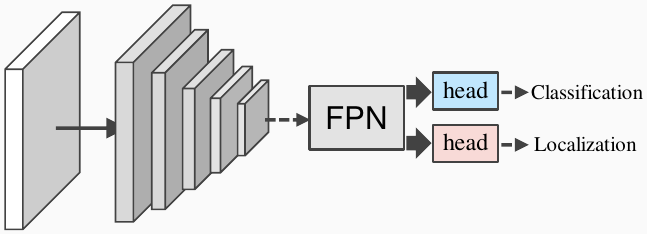}} 
  \caption{Ablation studies on variant disentanglement options. (a)-(d) indicate disentangling the detector from stride 8, stride 16, stride 32 and sibling head, respectively.} 
  \label{fig:subfig}
\end{figure}


\begin{table}[h]
\centering
\begin{center}
\begin{tabular}{c|c|c|c c c}
\hline  
Disentanglement & \#param &AP & AP$_{.5}$ & AP$_{.75}$\\
\hline
ResNet-50 & 41.8M& 36.1 &58.0 & 38.8\\
ResNet-50+D$_{s8}$ & 81.1M &22.3 & 46.3& 16.7\\
ResNet-50+ D$_{s16}$ &74.0M  &22.0 & 46.2 &16.3 \\
ResNet-50+ D$_{s32}$&59M  & 20.3 & 44.7 & 13.2\\
ResNet-50+ D$_{head}$&55.7M  & 37.3 & 59.4 & 40.2\\
\algname{} w/o \loss{}& 58.9M  &\bf{38.2} & \bf{60.5} & \bf{41.1}\\
\hline
\end{tabular}
\end{center}
\caption{Detailed performance and \#parameter of different disentanglement methods.}
\label{tab:ablation}
\end{table}

\textbf{Task-aware disentanglement.}
When it comes to tangled tasks conflict in sibling detection head, it's natural to think about decoupling different tasks from the backbone or detection head. To evaluate these ideas, we conduct several experiments to illustrate the comparison between them. As shown in Figure.\ref{fig:subfig}, we design different decoupling options including backbone disentanglement and head disentanglement. Detailed performance is shown in Table.\ref{tab:ablation}. Decoupling the classification and localization from the backbone largely degrades the performance. It clearly shows that the semantic information in the backbone should be shared by different tasks.
As expected, the task-specific head can significantly improve the performance.
Compared with D$_{head}$, \algname{} w/o \loss{} can further enhance the AP with the slight increased parameters, even for the demanding AP$_{.75}$.
When faced with heavy backbones, a slight increased parameter is trivial but can still significantly improve the performance.
This also substantiates the discussion in Sec.~\ref{discuss} that disentangling the tasks from spatial dimension can effectively alleviate the inherent conflict in sibling detection head.

\begin{table}[h]
\centering
\begin{center}
\scalebox{0.9}{
\begin{tabular}{c|c|c c c}
\hline  
Method & AP & AP$_{.5}$ & AP$_{.75}$\\
\hline
\algname{} w/o \loss{} & 38.2 &60.5 & 41.1\\
+ Joint training with sibling head $\mathcal{H}_{*}$& 39.7 & 61.7 & 42.8\\
\hline
\end{tabular}}
\end{center}
\caption{Result of joint training with sibling $\mathcal{H}_{*}$. The ResNet-50 with FPN is used as the basic detector.}
\label{tab:joint}
\end{table}

\textbf{Joint training with sibling head $\mathcal{H}_*$.}
In \algname{}, the shared proposal $P$ can also be used to perform classification and localization in an extra sibling head. We empirically observe that the training of sibling head is complementary to the training of \algname{}, and the results are demonstrated in Table.\ref{tab:joint}. 
This indicates that the derived proposals $\hat{P}_c$ and $\hat{P}_r$ are not conflict with the original proposal $P$. At the inference stage, only the \algname{} head is retained.

\begin{table}[h]
\centering
\begin{center}
\scalebox{0.9}{
\begin{tabular}{c|c| c| c| c | c| c}
\hline  
\multirow{2}{*}{Method} & \multirow{2}{*}{\algname{}} &\multicolumn{2}{|c|}{\loss{}} & \multirow{2}{*}{AP} & \multirow{2}{*}{AP$_{.5}$} & \multirow{2}{*}{AP$_{.75}$}\\
\cline{3-4}
 & & $\mathcal{M}_{cls}$ & $\mathcal{M}_{loc}$ & & & \\
\hline
 ResNet-50 & \checkmark &  &  & 39.7 & 61.7 & 42.8 \\
 ResNet-50 & \checkmark & \checkmark &  & 40.1&61.7	&43.2 \\
 ResNet-50 & \checkmark &  & \checkmark & 40.8 & 61.7 & 43.8 \\
 ResNet-50 & \checkmark & \checkmark & \checkmark & 41.0 & 61.7 & 44.3 \\
\hline
\end{tabular}}
\end{center}
\caption{Ablation studies on \loss{}. All of the experiments is joint training with sibling head $\mathcal{H}_{*}$. m$_c$ and m$_r$ are set to 0.2.}
\label{tab:PC}
\end{table}

\textbf{Effectiveness of \loss{}.}
In Sec.~\ref{L2}, we further propose the \loss{} to enhance the performance of \algname{}.
Table.\ref{tab:PC} reports the detailed ablations on it. We find that \loss{} significantly improves the AP$_{.75}$ by 1.5 and AP$_{.5}$ is barely affected.
This demonstrates that \loss{} aims to advocate more confidential classification and precise regression for the accurate boxes. 
Even on the strict testing standards AP (IoU from 0.5:0.95), 1.3 AP gain can also be obtained.

\begin{table}[h]
\centering
\begin{center}
\scalebox{0.9}{
\begin{tabular}{c c c c c c c}
\hline  
Method & $\emph{PC}$ & $\hat{P}_c$ & $\hat{P}_r$ & AP & AP$_{.5}$ & AP$_{.75}$\\
\hline
\algname{} & & $\emph{Point.w}$ & - & 38.0 & 60.3 & 40.89 \\
\algname{} & & $\emph{Point.w}$ & $\emph{Point.w}$ & 38.5 & 60.7& 41.7 \\
\algname{} & & $\emph{Point.w}$ & $\emph{Prop.w}$ & 38.2 & 60.5 & 41.1 \\
\hline
\algname{} &\checkmark & $\emph{Prop.w}$  & $\emph{Prop.w}$ & 39.8 & 60.1 & 42.9 \\
\algname{} &\checkmark & $\emph{Point.w}$ & $\emph{Point.w}$ & 40.7 & 61.8 & 44.4 \\
\algname{} &\checkmark & $\emph{Point.w}$ & $\emph{Prop.w}$ & 41.0 & 61.7 & 44.3 \\
\hline
\end{tabular}}
\end{center}
\caption{Results of different proposal learning manners for $\mathcal{H}^D_*$.}
\label{tab:deform}
\end{table}

\textbf{Derived proposal learning manner for $\mathcal{H}^D_*$.}
There are different programmable strategies to generate the derived proposal $\hat{P}_r$ and $\hat{P}_c$ including proposal-wise translation ($\emph{Prop.w}$) in Eq.~\ref{E4}, pointwise deformation ($\emph{Point.w}$) such as deformable RoI pooling~\cite{dai2017deformable} or the tricky combination of them.
To explore the differences of these learning manners, we conduct extensive experiments for 
COCO $\emph{minival}$ with ResNet-50.
Table.\ref{tab:deform} demonstrates the comparison results.
These comparisons illustrate that $\emph{Point.w}$ is beneficial to the classification task and cooperated with \loss{}, $\emph{Prop.w}$ performs a slight advantage on localization.
For generating the derived proposals, classification requires the optimal local features without regular shape restrictions and regression requires the maintenance of global geometric shape information.

\begin{figure}[h]
  \centering 
  \includegraphics[width=0.9\linewidth]{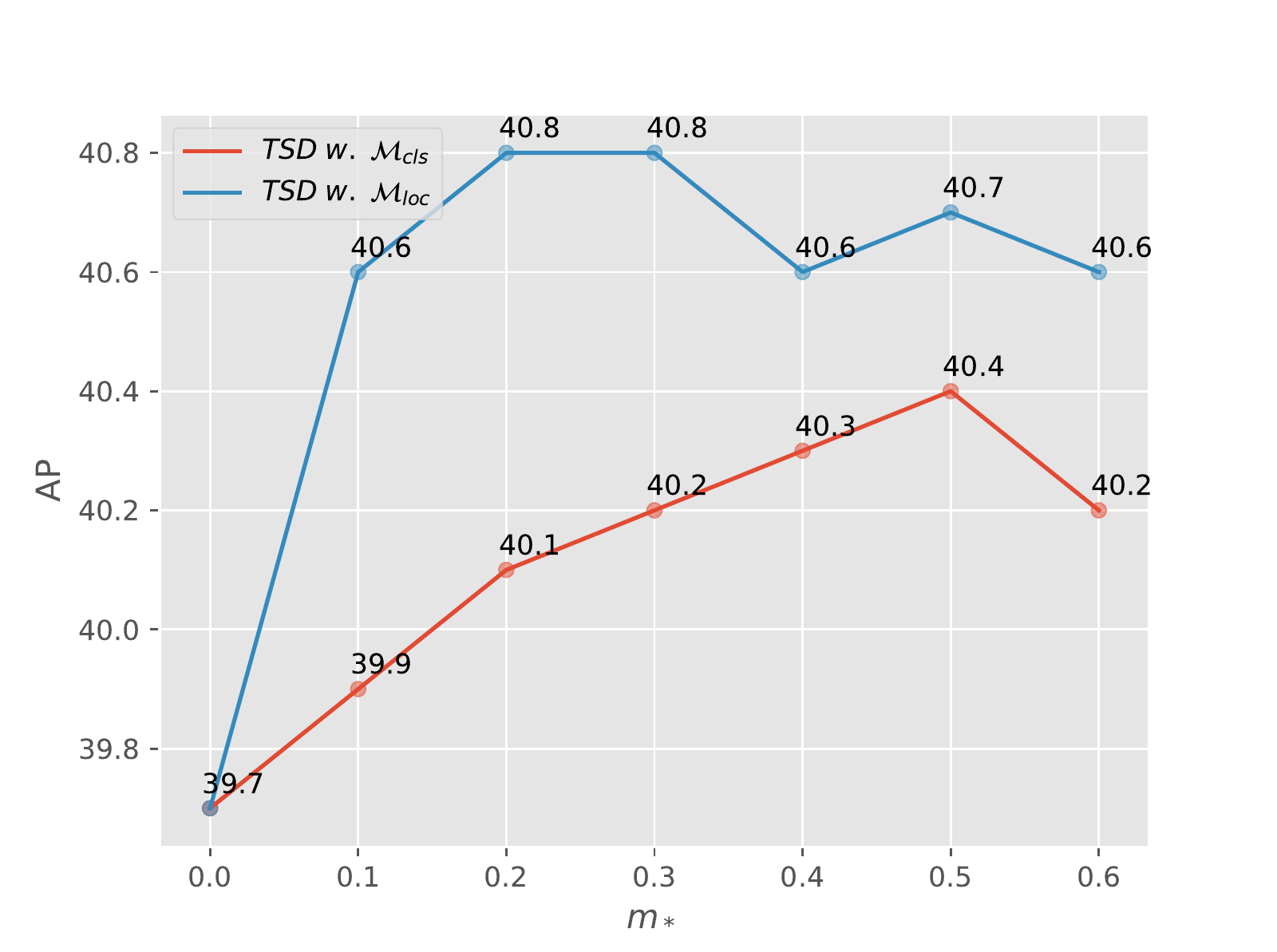}
  \caption{Results of \algname{} with variant m$_*$ for \loss{}. These experiments are conducted based on ResNet-50 with FPN.} 
  \label{fig:loss}
\end{figure}

\textbf{Delving to the effective \loss{}.}
\loss{} demonstrates its superiority on regressing more precise bounding boxes.
The hyper-parameters $m_c$ and $m_r$ play important roles in the training of \algname{} and to better understand their effects on performance, we conduct detailed ablation studies on them.
Figure.\ref{fig:loss} reports the results and note that both of the $\mathcal{M}_{los}$ and $\mathcal{M}_{cls}$ can further improve the performance.

\begin{table}[h]
\centering
\begin{center}
\scalebox{0.9}{
\begin{tabular}{c| c c c c c}
\hline  
Method & Ours & AP & AP$_{.5}$ & AP$_{.75}$ & runtime\\
\hline
ResNet-50& & 36.1 & 58.0 & 38.8 & 159.4 ms\\
ResNet-50& \checkmark & \bf{41.0} & \bf{61.7} & \bf{44.3} &174.9 ms\\
\hline
ResNet-101& & 38.6 & 60.6 & 41.8 & 172.4ms\\
ResNet-101& \checkmark & \bf{42.4} & \bf{63.1} & \bf{46.0} & 189.0ms\\
\hline
ResNet-101-DCN & & 40.8 & 63.2 & 44.6 & 179.3ms\\
ResNet-101-DCN & \checkmark &\bf{43.5} & \bf{64.4} & \bf{47.0} & 200.8ms\\
\hline
ResNet-152& & 40.7 & 62.6 & 44.6 & 191.3ms\\
ResNet-152&\checkmark & \bf{43.9} & \bf{64.5} & \bf{47.7} & 213.2ms\\
\hline
ResNeXt-101~\cite{xie2017aggregated}& &  40.5& 62.6 & 44.2 & 187.5ms \\
ResNeXt-101~\cite{xie2017aggregated}& \checkmark & \bf{43.5} & \bf{64.5} & \bf{46.9} & 206.6ms\\
\hline

\end{tabular}}
\end{center}
\caption{Results of \algname{} + \loss{} with variant backbones. DCN means deformable convolution. The runtime includes network forward and post-processing (e.g., NMS for object detection). The runtime is the averaged value on a single Tesla V100 GPU and CPU E5-2680 v4.}
\label{tab:backbones}
\end{table}

\subsection{Applicable to variant backbones}
Since the \algname{} and \loss{} have demonstrated their outstanding performance on ResNet-50 with FPN, we further delve into the adaptation on variant backbones.
Based on Faster R-CNN, we directly conduct several experiments with different backbones and Table.\ref{tab:backbones} summarizes the detailed performance on COCO $\emph{minival}$.
TSD can steadily improve the performance by 3\%$\sim$5\% with additional $\sim$10\% time cost. Note that ResNet-50+TSD with 58.9M parameter can even outperform the ResNet-152 with 76.39M parameter.
Based on the ResNet family, TSD is a more preferred choice than increasing backbone to improve performance.
\textbf{If not specified, all subsequent \algname{} indicates \algname{}+\loss{}}.


\begin{table}[h]
\centering
\begin{center}
\scalebox{.9}{
\begin{tabular}{c| c c c c }
\hline  
Method & \algname{} & AP$_{.5}$ (Val) & AP$_{.5}$ (LB)\\
\hline
ResNet-50& & 64.64 & 49.79 \\
ResNet-50&\checkmark& \bf{68.18} & \bf{52.55} \\
\hline
Cascade-DCN-SENet154 & & 69.27 & 55.979 \\
Cascade-DCN-SENet154 &\checkmark & \bf{71.17} & \bf{58.34}\\
\hline
DCN-ResNeXt101$^*$ & & 68.70 & 55.05 \\
DCN-ResNeXt101$^*$ &\checkmark & \bf{71.71} & \bf{58.59} \\
\hline
DCN-SENet154$^*$ & & 70 &57.771 \\
DCN-SENet154$^*$ &\checkmark & \bf{72.19} & \bf{60.5}\\
\hline
\end{tabular} }
\end{center}
\caption{Results of \algname{} on OpenImage dataset. * indicates we expand the anchor scale to \{8, 11, 14\} and anchor aspect ratio to \{0.1, 0.5, 1, 2, 4, 8\}. Furthermore, mult-scale test is used for public leaderboard (LB) except for ResNet-50.}
\label{tab:openimage}
\end{table}

\begin{table*}[h]
\centering
\begin{center}
\scalebox{0.9}{
\begin{tabular}{c| c c c c c c c}
\hline  
Method & Ours & AP$^{bb}$ & AP$^{bb}_{.5}$ & AP$^{bb}_{.75}$ & AP$^{mask}$ & AP$^{mask}_{.5}$& AP$^{mask}_{.75}$\\
\hline
ResNet-50 w. FPN& & 37.2 &58.8& 40.2& 33.6& 55.3& 35.4 \\
ResNet-50 w. FPN&\checkmark & \bf{41.5} & \bf{62.1} & \bf{44.8} & \bf{35.8} & \bf{58.3}& \bf{37.7} \\
\hline
ResNet-101 w. FPN & & 39.5& 61.2 & 43.0 & 35.7 & 57.9 & 38.0 \\
ResNet-101 w. FPN &\checkmark & \bf{43.0}& \bf{63.6} & \bf{46.8} & \bf{37.2} & \bf{59.9} & \bf{39.5} \\
\hline
\end{tabular}}
\end{center}
\caption{Results of Mask R-CNN with \algname{}. The proposed methods are only applied on the detection branch in Mask R-CNN. AP$^{bb}$ means the detection performance and AP$^{mask}$ indicates the segmentation performance.}
\label{tab:mask}
\end{table*}

\begin{table*}[t!]
\centering
\begin{center}
\scalebox{0.9}{
\begin{tabular}{c| c | c |c|c c c c c}
\hline  
Method & backbone & $b\&w$ &AP  & AP$_{.5}$ & AP$_{.75}$ & AP$_s$ & AP$_m$ & AP$_l$ \\
\hline
RefineDet512~\cite{zhang2018single} & ResNet-101& & 36.4 & 57.5 & 39.5& 16.6 & 39.9 & 51.4 \\
RetinaNet800~\cite{lin2017focal} & ResNet-101& & 39.1 & 59.1 & 42.3 & 21.8 &42.7 & 50.2 \\
CornerNet~\cite{law2018cornernet} & Hourglass-104~\cite{newell2016stacked}& & 40.5 & 56.5 &43.1& 19.4 &42.7 & 53.9 \\
ExtremeNet~\cite{zhou2019bottom}& Hourglass-104~\cite{newell2016stacked}& & 40.1 & 55.3 & 43.2& 20.3 &43.2 & 53.1 \\
FCOS~\cite{Tian_2019_ICCV} & ResNet-101& & 41.5& 60.7& 45.0 & 24.4 & 44.8 &51.6 \\
RPDet~\cite{yang2019reppoints} & ResNet-101-DCN& \checkmark & 46.5 & 67.4 & 50.9 & 30.3& 49.7& 57.1 \\
CenterNet511~\cite{Duan_2019_ICCV} & Hourglass-104& \checkmark& 47.0 &64.5& 50.7& 28.9& 49.9& 58.9 \\
TridentNet~\cite{li2019scale}& ResNet-101-DCN& \checkmark & 48.4 & 69.7& 53.5 & 31.8& 51.3& 60.3\\
NAS-FPN~\cite{ghiasi2019fpn}& AmoebaNet (7 @ 384)& \checkmark & 48.3 & -& - &- &-&- \\
\hline
Faster R-CNN w FPN~\cite{lin2017feature} & ResNet-101& &36.2& 59.1& 39.0& 18.2& 39.0& 48.2\\
Auto-FPN$^{\dagger}$~\cite{Xu_2019_ICCV}& ResNet-101& & 42.5& - &-& -& -& -\\
Regionlets~\cite{xu2018deep}& ResNet-101& &39.3& 59.8& - & 21.7& 43.7& 50.9 \\
Grid R-CNN~\cite{lu2019grid}& ResNet-101& & 41.5& 60.9& 44.5& 23.3& 44.9&54.1 \\
Cascade R-CNN~\cite{cai2018cascade}& ResNet-101& & 42.8 & 62.1& 46.3 & 23.7& 45.5& 55.2\\
DCR~\cite{Cheng_2018_ECCV}& ResNet-101& &40.7& 64.4 &44.6 &24.3& 43.7& 51.9 \\

IoU-Net$^{\dagger}$~\cite{jiang2018acquisition} & ResNet-101 & & 40.6 & 59.0& -& - & - & - \\
Double-Head-Ext$^{\dagger}$~\cite{wu2019rethinking} & ResNet-101 & & 41.9 & 62.4 &45.9 &23.9&45.2&55.8\\

SNIPER~\cite{singh2018sniper}& ResNet-101-DCN& \checkmark& 46.1& 67.0&51.6& 29.6& 48.9& 58.1\\
DCNV2~\cite{zhu2019deformable} & ResNet-101& \checkmark &46.0& 67.9& 50.8& 27.8& 49.1& 59.5\\
PANet~\cite{liu2018path} & ResNet-101& \checkmark &47.4& 67.2&51.8 &30.1 &51.7&60.0 \\
GCNet~\cite{cao2019gcnet}& ResNet-101-DCN&\checkmark & 48.4&67.6&52.7&-&-&-\\
\hline
\bf{\algname{}}$^{\dagger}$ &ResNet-101 & &\bf{43.1} &\bf{63.6}&\bf{46.7}&\bf{24.9}&\bf{46.8}& \bf{57.5}\\
\bf{\algname{}}& ResNet-101 & & \bf{43.2} &{64.0} & \bf{46.9}& \bf{24.0}& \bf{46.3}& \bf{55.8} \\
\bf{\algname{}}$^*$& ResNet-101-DCN &\checkmark & \bf{49.4} &{69.6} & \bf{54.4}& \bf{32.7}& \bf{52.5}& \bf{61.0} \\
\bf{\algname{}}$^*$& SENet154-DCN~\cite{hu2018squeeze} & \checkmark &\bf{51.2}&\bf{71.9}&\bf{56.0}&\bf{33.8}&\bf{54.8}&\bf{64.2} \\
\hline
\end{tabular}}
\end{center}
\caption{Comparisons of single-model results for different algorithms evaluated on the COCO $\emph{test-dev}$ set. $b\&w$ indicates training with bells and whistles such as multi-scale train/test, Cascade R-CNN or DropBlock~\cite{ghiasi2018dropblock}. $\dagger$ indicates the result on COCO $\emph{minival}$ set.}
\label{tab:SOTA}
\end{table*}

\subsection{Applicable to Mask R-CNN}
The proposed algorithms largely surpass the classical sibling head in Faster R-CNN. Its inherent properties determine its applicability to other R-CNN families such as Mask R-CNN for instance segmentation. To validate this, we conduct experiments with Mask R-CNN~\cite{he2017mask}. Performances are shown in Table.\ref{tab:mask} and the training configuration in Mask R-CNN is the same as the experiments in Faster R-CNN.
It's obvious that \algname{} is still capable of detection branch in Mask R-CNN.
The instance segmentation mask AP can also obtain promotion.

\subsection{Generalization on large-scale \bf{OpenImage}}
In addition to evaluate on the COCO dataset, we further corroborate the proposed method on the large-scale OpenImage dataset.
As the public dataset with large-scale boxes and hierarchy property, it brings a new challenge to the generalization of detection algorithms. To fully delve the effectiveness of the proposed algorithm, we run a number of ablations to analyze \algname{}.
Table.\ref{tab:openimage} illustrates the comparison and note that, even for heavy backbone, \algname{} can still give satisfactory improvements. 
Furthermore, \algname{} is complementary to Cascade R-CNN~\cite{cai2018cascade} and embedding it into this framework can also enhance the performance by a satisfactory margin.

\begin{figure*}[h]
  \centering 
  \includegraphics[width=0.8\linewidth]{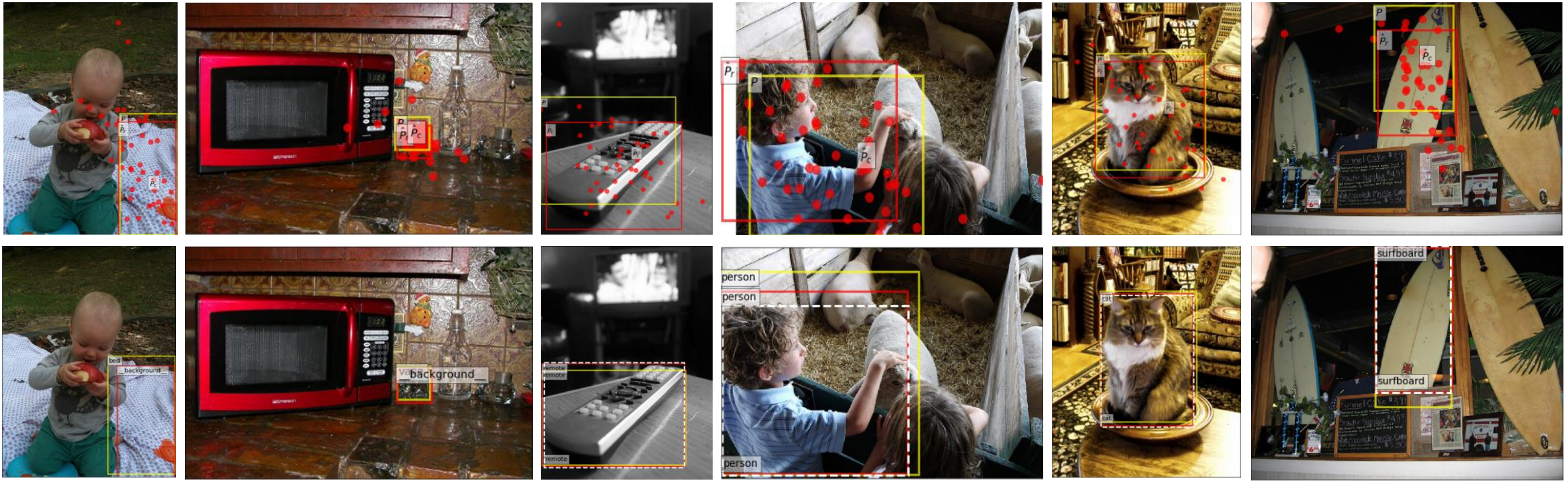}
  \caption{Visualization of the learnt $\hat{P}_r$ and $\hat{P}_c$ on examples from the COCO \emph{minival} set. The first row indicates the proposal $P$ (yellow box) and the derived $\hat{P}_r$ (red box) and $\hat{P}_c$ (pink point, center point in each grid). The second row is the final detected boxes where the white box is ground-truth. \algname{} deposes the false positives in the first two columns and in other columns, it regresses more precise boxes.} 
  \label{fig:vis}
\end{figure*}

\begin{figure}[h]
  \centering 
  \includegraphics[width=0.8\linewidth]{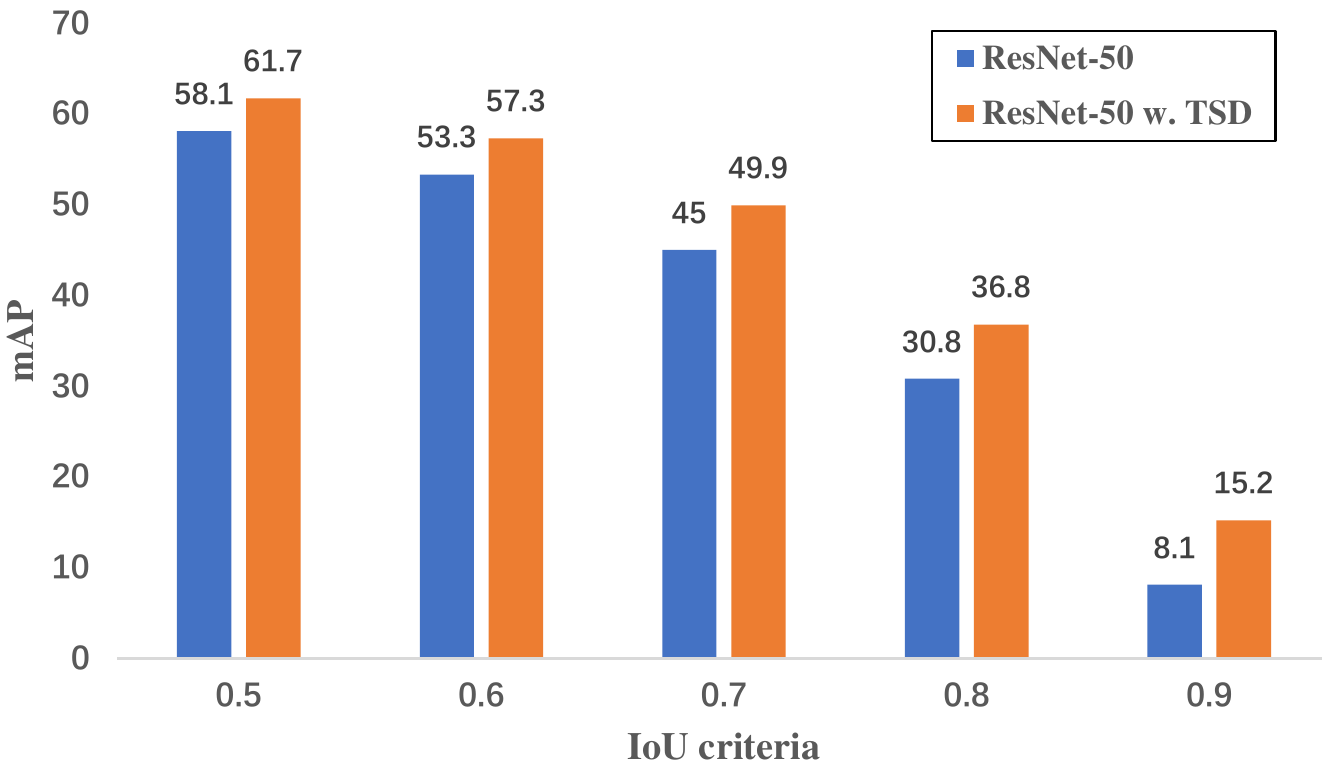}
  \caption{mAP across IoU criteria from 0.5 to 0.9 with 0.1
interval.} 
  \label{fig:iou}
\end{figure}

\subsection{Comparison with state-of-the-Arts}
In this section, we evaluate our proposed method on COCO $\emph{test-dev}$ set and compare it with other state-of-the-art methods. $m_c$ and $m_r$ are set to 0.5 and 0.2, respectively.
For a fair comparison, we report the results of our methods under different settings in Table.\ref{tab:SOTA}.
For comparison with Grid R-CNN~\cite{lu2019grid}, we extend the training epochs for ResNet-101 to be consistent with it.
For comparing with the best single-model TridentNet$^*$, in \textbf{\algname{}}$^*$, we apply the same configuration with it including multi-scale training, soft-NMS~\cite{bodla2017soft}, deformable convolutions and the $3\times$ training scheme on ResNet-101.
The best single-model ResNet-101-DCN gives an \textbf{AP of 49.4}, already surpassing all of the other methods with the same backbone. To our best knowledge, for a single model with ResNet-101 backbone, our result is the best entry among the state-of-the-arts.
\algname{} demonstrates its advantage on promoting precise localization and confidential classification, especially on higher IoU thresholds (AP$_{.75}$).
Furthermore, we explore the upper-bound of \algname{} with a heavy backbone.
Surprisingly, it can achieve the \textbf{AP of 51.2} with the single-model SENet154-DCN on COCO $test$-$dev$ set. Soft-NMS is not used in this evaluation.

\subsection{Analysis and discussion}
\textbf{Performance in different IoU criteria}.
Since \algname{} exhibits superior ability on regressing precise localization and predicting confidential category, we conduct several evaluations with more strict IoU criteria on COCO $minival$.
Figure.\ref{fig:iou} illustrates the comparison
between \algname{} based Faster R-CNN and baseline Faster R-CNN with
the same ResNet-50 backbone across IoU thresholds from
0.5 to 0.9.
Obviously, with the increasing IoU threshold, the improvement brought by \algname{} is also increasing.

\textbf{Performance in different scale criteria}.
We have analyzed the effectiveness of \algname{} under different IoU criteria.
To better explore the specific improvement, we further test the mAP under objects with different scales. Table.\ref{tab:scale} reports the performance and \algname{} shows successes in objects with variant scales, especially for medium and large objects.

\textbf{What did \algname{} learn?}
Thanks to the \algfullname{} (\algname{}) and the progressive constraint (\loss{}),
stable improvements can be easily achieved whether for variant backbones or variant datasets.
Beyond the quantitative promotion, we wonder what \algname{} learned compared with the sibling head in Faster R-CNN.
To better interpret this, We showcase the illustrations of our \algname{} compared with sibling head as shown in Figure.~\ref{fig:vis}.
As expected, through \algname{}, it can depose many false positives and regress the more precise box boundary. For $\hat{P}_r$, it tends to translate to the boundary that is not easily regressed.
For $\hat{P}_c$, it tends to concentrate on the local appearance and object context information as it did in sibling head with deformable RoI pooling~\cite{dai2017deformable}. Note that the tangled tasks in sibling head can be effectively separated from the spatial dimension.

\begin{table}[t]
\centering
\begin{center}
\scalebox{0.9}{
\begin{tabular}{c| c| c c c c c}
\hline  
Criteria& \algname{} &AP$_{.5}$ & AP$_{.6}$& AP$_{.7}$ & AP$_{.8}$ & AP$_{.9}$\\
\hline
AP$_{small}$& &38.4 & 33.7 & 26.7 & 16.2& 3.6 \\
AP$_{small}$& \checkmark & \bf{40.0}&\bf{35.6}&\bf{28.8}&\bf{17.7}&\bf{5.3}\\
\hline
AP$_{medium}$& & 62.9&58.4 &49.7&33.6&8.7 \\
AP$_{medium}$&\checkmark & \bf{67.7}&\bf{62.4} &\bf{54.9} &\bf{40.2} &\bf{15.4} \\
\hline
AP$_{large}$ & &69.5 &65.5&56.8 & 43.2 &14.8 \\
AP$_{large}$ & \checkmark &\bf{74.8} & \bf{71.6}&\bf{65.0} &\bf{53.2} &\bf{27.9}\\
\hline
\end{tabular}}
\end{center}
\caption{mAP across scale criteria from 0.5 to 0.9 with 0.1 interval.}
\label{tab:scale}
\end{table}

\section{Conclusion}
In this paper, we present a simple operator \algname{} to alleviate the inherent conflict in sibling head, which learns the task-aware spatial disentanglement to bread through the performance limitation.
In particular, \algname{} derives two disentangled proposals from the shared proposal and learn the specific feature representation for classification and localization, respectively. Further, we propose a progressive constraint to enlarge the performance margin between the disentangled and the shared proposals, which provides additional performance gain.
Without bells and whistles, this simple design can easily boost most of the backbones and models on both COCO and large scale OpenImage consistently by 3\%$\sim$5\%, and is the core model in our 1st solution of OpenImage Challenge 2019.

{\small
\bibliographystyle{ieee_fullname}
\bibliography{egbib}
}

\end{document}